\pdfoutput=1

\documentclass[11pt]{article}
\usepackage{graphicx}
\usepackage{amsmath,amsfonts,bm}
\usepackage{array}
\graphicspath{{./Figures/}}
\usepackage{ACL2023}

\usepackage{times}
\usepackage{latexsym}
\usepackage{hyperref}
\usepackage{url}
\usepackage{bbm}
\usepackage{amsmath,amsfonts,bm}
\usepackage{graphicx}
\usepackage{multirow}
\usepackage{makecell}
\usepackage{array}
\usepackage[para]{threeparttable}
\usepackage{booktabs}
\usepackage{tabularx}
\usepackage{xspace}
\usepackage{subcaption}
\usepackage{multirow}
\usepackage[normalem]{ulem}
\usepackage[T1]{fontenc}

\usepackage[utf8]{inputenc}

\usepackage{microtype}

\usepackage{inconsolata}
\usepackage{xcolor}

\newcommand{\method}{GAP\xspace}
\title{Gradient Ascent Post-training Enhances Language Model Generalization}

\author{Dongkeun Yoon{\textsuperscript{1\thanks{\;\;Equal Contribution}}}\quad Joel Jang{\textsuperscript{1\footnotemark[1]}} \quad Sungdong Kim{\textsuperscript{1, 2}} \quad Minjoon Seo{\textsuperscript{1}} \\
{\textsuperscript{1}}KAIST \quad{\textsuperscript{2}} NAVER AI Lab \\
\texttt{mattyoon99@gmail.com, \{joeljang,minjoon\}@kaist.ac.kr} \\ \texttt{sungdong.kim@navercorp.com} \\
}

\begin{document}
\maketitle
\begin{abstract}
In this work, we empirically show that updating pretrained LMs (350M, 1.3B, 2.7B) with just a few steps of \textbf{G}radient \textbf{A}scent \textbf{P}ost-training (\method) on random, unlabeled text corpora enhances its zero-shot generalization capabilities across diverse NLP tasks. Specifically, we show that GAP can allow LMs to become comparable to 2-3x times larger LMs across 12 different NLP tasks. We also show that applying \method on out-of-distribution corpora leads to the most reliable performance improvements. Our findings indicate that \method can be a promising method for improving the generalization capability of LMs without any task-specific fine-tuning~\footnote{Code and full results for individual GAP runs are available at \href{https://github.com/kaist-lklab/GAP}{https://github.com/kaist-lklab/GAP}}.
\end{abstract}

\section{Introduction}
\label{sec:intro}
Recently, Language Models (LMs) pretrained on a vast amount of text corpora have shown to be capable of performing diverse downstream NLP tasks in a zero-shot manner~\citep{brown2020language, rae2021scaling, chowdhery2022palm, zhang2022opt} or through in-context learning~\citep{brown2020language, min2022rethinking} without any gradient updates. This paradigm has been preferred over task-specific fine-tuning~\citep{devlin-etal-2019-bert}, which requires considerable amount of labeled data for the given target task. 

\begin{figure}[t!]
\includegraphics[width=0.75\linewidth]{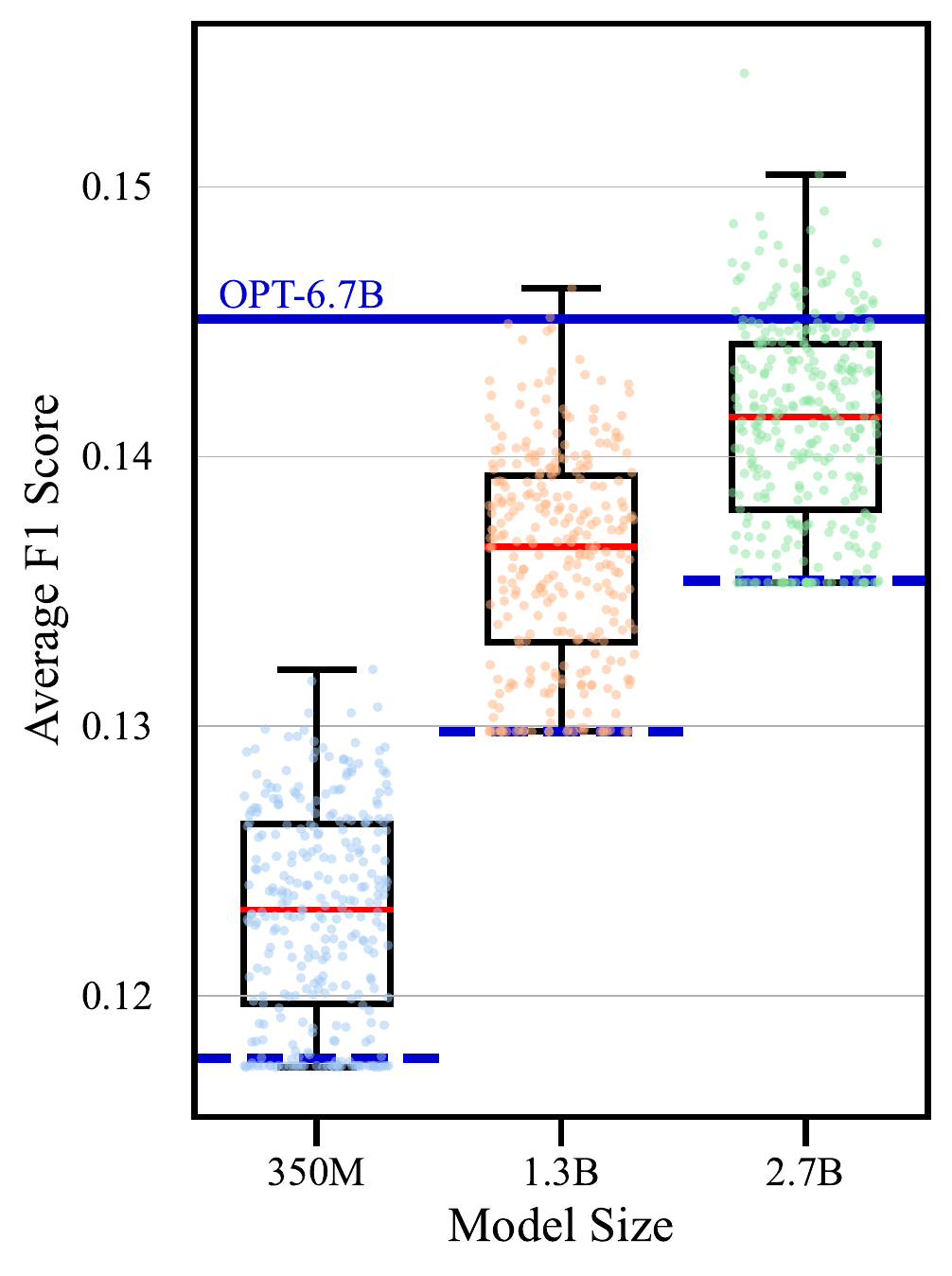}
\centering
\caption{Average validation F1-score measured on four dialogue datasets. A single dot represents a single GAP run, each with random text samples (total of 300 runs  per LM size). The dashed horizontal lines indicate performance of OPT LMs (baseline) of the same size. For reference we also show the performance of 6.7B-OPT baseline with a solid line.}
\label{fig:dialog_val_f1}
\vspace{-3mm}
\end{figure}

Motivated by the positive effect of gradient ascent during fine-tuning~\citep{foret2021sharpnessaware}, in this work, we explore whether adapting pretrained LMs with \textbf{G}radient \textbf{A}scent \textbf{P}ost-training (\method) on random, unlabeled text corpora can bring any benefits in terms of enhancing its generalization capabilities of performing diverse downstream NLP tasks in a zero-shot or few-shot manner \textit{without} the need for task-specific training data. 


Specifically, we apply just a few steps of gradient ascent to OPT LMs~\cite{zhang2022opt} using randomly sampled text sequences from 3 different corpora from the Pile~\cite{gao2021pile} with varying degree of familiarity between the LM and the corpus. Experimental results show that this simple approach achieves performance gains across 12 downstream NLP tasks: 
4 dialogue tasks and 8 classification tasks.
We observe that applying \method with out-of-distribution data, specifically code data that OPT was not explicitly trained on, results in the most reliable performance gain.

Our main contributions can be summarized into two folds:
\begin{itemize}
    \item We empirically show that \method is a promising generalization enhancement technique as it is (1) effective, as evidenced by multiple benchmark results; (2) simple \& efficient, requiring maximum 15 steps of parameter update; (3) versatile, as it can be applied easily to any pretrained LMs and does not necessitate task-specific fine-tuning.
    
    \item We show analysis of what makes \method work by splitting the corpora into three groups according to the LMs' degree of familiarity with the data. We observe that performing \method with the most unfamiliar (out-of-distribution) data results in the most reliable performance gain.
    
\end{itemize}

\section{Related Works}
\paragraph{Task-Specific Gradient Ascent}
Deep neural network models exhibiting poor generalization due to converging at sharp local minima is a well-known phenomenon in literature~\cite{keskar2017on, Izmailov2018SWA, Cha2021SWAD, chen2022when}. To address this issue, \citet{foret2021sharpnessaware} introduce Sharpness-Aware Minimization (SAM), an algorithm that performs both gradient ascent as well as gradient descent during task-specific fine-tuning to avoid sharp local minima, improving performance. The effectiveness of SAM has motivated several studies to apply them to LMs and report meaningful improvements in performance.

\citet{bahri2022sharpness} have shown that applying SAM when fine-tuning various scales of T5 LMs~\citep{2020t5} on multiple downstream tasks results in a substantial performance gains. Similarly, \citet{kaddour2022flat} also explore SAM across computer vision, natural language processing, and graph representation learning tasks, further bolstering its efficiency. 

While SAM was proposed as a robust fine-tuning methodology that targets convergence on supervised dataset, we instead explore the benefits gradient ascent can bring \textit{without} task-specific labeled data for generic LMs.

\paragraph{Task-Agnostic Gradient Ascent}
In a recent study, \citet{jang-knowledge_unlearning-2022} investigate the use of gradient ascent for addressing privacy risks in LMs. The main objective of the work is utilizing gradient ascent to \textit{unlearn} specific token sequences; surprisingly, they report unexpected performance gains in some cases. Our work can be seen as a direct extension of this phenomenon where our main objective is to enhance the generalization capabilities instead of forgetting specific data to ensure privacy.

\section{Gradient Ascent Post-training (\method)}
In this section, we give a formal definition of \method. Specifically, given an LM with parameters $w$ and a sequence of tokens $\boldsymbol{x}=(x_1,...,x_N)$, \method is defined as:

\begin{equation}
w_{t+1} = w_t + \alpha\nabla f_{w_t}(\boldsymbol{x})\label{eq:1}
\end{equation}
\begin{equation}
f_{w_t}(\boldsymbol{x}) = - \sum_{n=1}^{N}\text{log}(p_{w_t}(x_{n}|x_{<n}))\label{eq:2}
\end{equation}
where $t$ represents the gradient ascent iteration, $\alpha$ denotes the learning rate, $x_{<n}$ indicates the token sequence $(x_1,...,x_{n-1})$ and $p_{w_t}(x_{n}|x_{<n})$ represents the likelihood of predicting the next token, $x_n$, given the previous token sequence as an input to an LM with parameter $w_t$.

Markedly, GAP solely utilizes gradient ascent and does not actively facilitate convergence, as it updates the model parameters to maximize (\ref{eq:1}) the language modeling loss function (\ref{eq:2}). We propose GAP as an unsupervised methodology that can bring significant performance gains even without curated fine-tuning data.


\section{Experiments}
\label{sec:experiments}
\subsection{Experimental Setup}
\paragraph{Baseline Models and Evaluation Datasets}
We use OPT (350M, 1.3B, 2.7B, 6.7B) LMs~\cite{zhang2022opt} as the baseline LMs. We observe the effect \method has on their generalization capabilities which is measured via evaluation on 12 different downstream NLP tasks; we use Wizard of Wikipedia~\cite{dinan2018wizard}, Empathetic Dialogues~\cite{rashkin-etal-2019-towards}, Blended Skill Talk~\cite{smith-etal-2020-put} and WizInt~\cite{komeili-etal-2022-internet} to evaluate generative capabilities, Hellaswag~\cite{zellers2019hellaswag} to assess linguistic reasoning abilities, Winogrande~\cite{sakaguchi2021winogrande} and COPA~\cite{brassard-etal-2022-copa} to measure commonsense reasoning abilities, and ARC-Easy~\cite{Clark2018ThinkYH}, ARC-Challenge~\cite{Clark2018ThinkYH}, PIQA~\cite{bisk2020piqa}, MathQA~\cite{amini-etal-2019-mathqa} and PubmedQA~\cite{jin2019pubmedqa} to measure the scientific reasoning abilities.
The exact prompts used for each task are provided in Appendix \ref{appen:task_prompt}.


\paragraph{Random Unlabeled Data}
We apply \method on text snippets from three different corpora, which all originate from the Pile~\cite{gao2021pile} training set: (1) Training Data Extraction Challenge (TDEC)\footnote{https://github.com/google-research/lm-extraction-benchmark}, (2) Common Crawl (CC) and (3) Github (Git.). We choose these corpora in order to observe the effect of the LMs' degree of familiarity with the data. Training Data Extraction Challenge includes examples from the Pile that are identified to be easy-to-extract from GPT-Neo LMs~\cite{black-etal-2022-gpt}, mainly due to high levels of duplication. We assume these examples are also relatively easier-to-extract from OPT LMs as they were also pretrained on subset of the Pile, indicating the highest level of familiarity / memorization. We consider OPT LMs to be familiar (in-domain) to Common Crawl, as it was included in their pretraining corpora. As OPT LMs were not explicitly trained on the Github corpora we consider OPT to be unfamiliar (out-of-distribution) with Github. Examples of the random unlabeled data are provided in Appendix \ref{appen:random_data}.

\paragraph{Configurations}
For each of the 3 LM sizes [350M, 1.3B, 2.7B], we sample a total of 300 text samples (each 200 token lengths long) for applying \method, with 100 samples taken from each of the three corpora. For each run, a single text sample is used, ultimately resulting in 300 runs of GAP per LM size. Therefore, a single epoch of a GAP run comprises of a single gradient ascent step with batch size set to 1. The number of maximum epochs is set to 15 and we report the validation score from the best-performing epoch, as preliminary experiments showed gradient ascent past 15 steps mostly resulted in performance degradation. Due to computational constraints we sample the validation data to a maximum of 320 samples per dataset for all of the 12 evaluation datasets. For further exploration of GAP as a methodology, we use the checkpoints with the best validation scores and evaluate the LMs on the test datasets for the 4 dialogue tasks. We do not separately report the test evaluation results for classification datasets since most of them require direct submission to the task website. For a single run, we use one Nvidia 40GB A100 GPU. Further details regarding the experimental configurations (e.g. optimizer, learning rate, etc.) are provided in Appendix \ref{appen:configs}.

\begin{table}[t]
\centering
\begin{tabularx}{0.485\textwidth}{l|cccc}
\toprule
\textbf{Model} & \textbf{F1} & \textbf{MAUVE} & \textbf{Diversity} & \textbf{Length}\\
\midrule
350M & 11.4 & 44.3 & 74.0 & 11.8 \\
\textit{+ GAP} & \textbf{12.5} & \textbf{67.2} & \textbf{87.3} & 14.4 \\
\midrule
1.3B & 13.5 & 48.2 & 82.8 & 11.4 \\
\textit{+ GAP} & \textbf{14.0} & \textbf{69.5} & \textbf{86.7} & 13.8 \\
\midrule
2.7B & 13.8 & 51.3 & 86.9 & 11.3 \\
\textit{+ GAP} & \textbf{14.7} & \textbf{73.0} & \textbf{93.1} & 14.5 \\
\midrule
6.7B & 14.5 & 51.1 & 88.3 & 11.9 \\
\bottomrule
\end{tabularx}
\caption{
Average test scores on dialogue datasets. We evaluate OPT baselines and our best-performing checkpoints excluding outliers. Individual results are provided in Appendix \ref{appen:full_results}.}
\label{table:test_dialog}
\end{table}

\subsection{Dialogue Tasks}
\paragraph{Main Results} 
As shown in Figure \ref{fig:dialog_val_f1} in Section \ref{sec:intro}, \method substantially enhances the average validation performance on the 4 dialogue tasks, with median F1-score of 1.3B LMs outperforming the 2.7B LM baseline, and some 1.3B LMs even able to match the performance of the 6.7B LM baseline~\footnote{Detailed numerical data for the median values is available in \ref{appen:full_results}.}. 
We report the average test F1 score as well as MAUVE~\citep{pillutla-etal:mauve:neurips2021}, diversity~\citep{su2022a}, and generation length of our best validation checkpoints for each model size (excluding outliers) in comparison to the baseline LMs in Table \ref{table:test_dialog}~\footnote{Explanation of how MAUVE and diversity is measured is provided in Appendix \ref{appen:configs}.}.

Results show a substantial improvement in all of the metrics, F1 Score, MAUVE, and generation length, with our 1.3B and 2.7B LM checkpoints even outperforming the larger LM baselines. This result is significant considering that no task-specific dataset is used.
Examples of text generation for the dialogue tasks are provided in Appendix \ref{appen:examples}.


\begin{table}[t]
\centering
\setlength{\tabcolsep}{4pt}
\begin{tabular}{c|c|ccc}
\hline
\textbf{Comparison} & \textbf{Metric} & \textbf{Win} & \textbf{Loss} & \textbf{Tie}\\
 \toprule
\multirow{3}{*}{Ours vs. Baseline} & C & \textbf{43}\%$^{\dagger}$ & 17\% & 40\% \\
& F & \textbf{36}\%$^{\dagger}$ & 15\% & 49\% \\
& I & \textbf{40}\%$^{\dagger}$ & 17\% & 43\% \\
\midrule
\multirow{3}{*}{Ours vs. Human} & C & \textbf{41}\% & 37\% & 22\% \\
& F & \textbf{33}\% & 30\% & 37\% \\
& I & 23\% & \textbf{50}\%$^{\dagger}$ & 27\% \\
\bottomrule
\end{tabular}
\caption{
Human evaluation results from dialogue generation task, WizInt~\cite{komeili-etal-2022-internet}. The C, F, and I indicate coherence, fluency, and informativeness, respectively. $\dagger$ indicates the significance with p-value lower than $0.1$ by bootstrap test between pairs.}
\label{table:human_eval}
\end{table}

\paragraph{Human Evaluation}
We also evaluate and compare the qualitative quality of generated responses of the baseline LMs and the LMs adapted with GAP side-by-side. For this, we sample 100 contexts from the WizInt~\cite{komeili-etal-2022-internet} dataset and generate the corresponding responses with the 2.7B LM baseline and 2.7B LM + \method denoted as \textit{Ours}. Then, we compare the generated response pairs from the LMs from the perspective of three metrics: coherence, fluency, and informativeness~\cite{su2022a}. We ask human evaluators to select the better response from each pair with respect to each metrics~\footnote{Further study details are in Appendix~\ref{sec:human_eval_detail}.}. We find our GAP-enhanced LM shows significant strengths in all the metrics compared to its baseline (Table~\ref{table:human_eval}). Moreover, our LM shows comparable performance to human upper bounds (gold response) except for informativeness.

\begin{figure}[t!]
\includegraphics[width=0.75\linewidth]{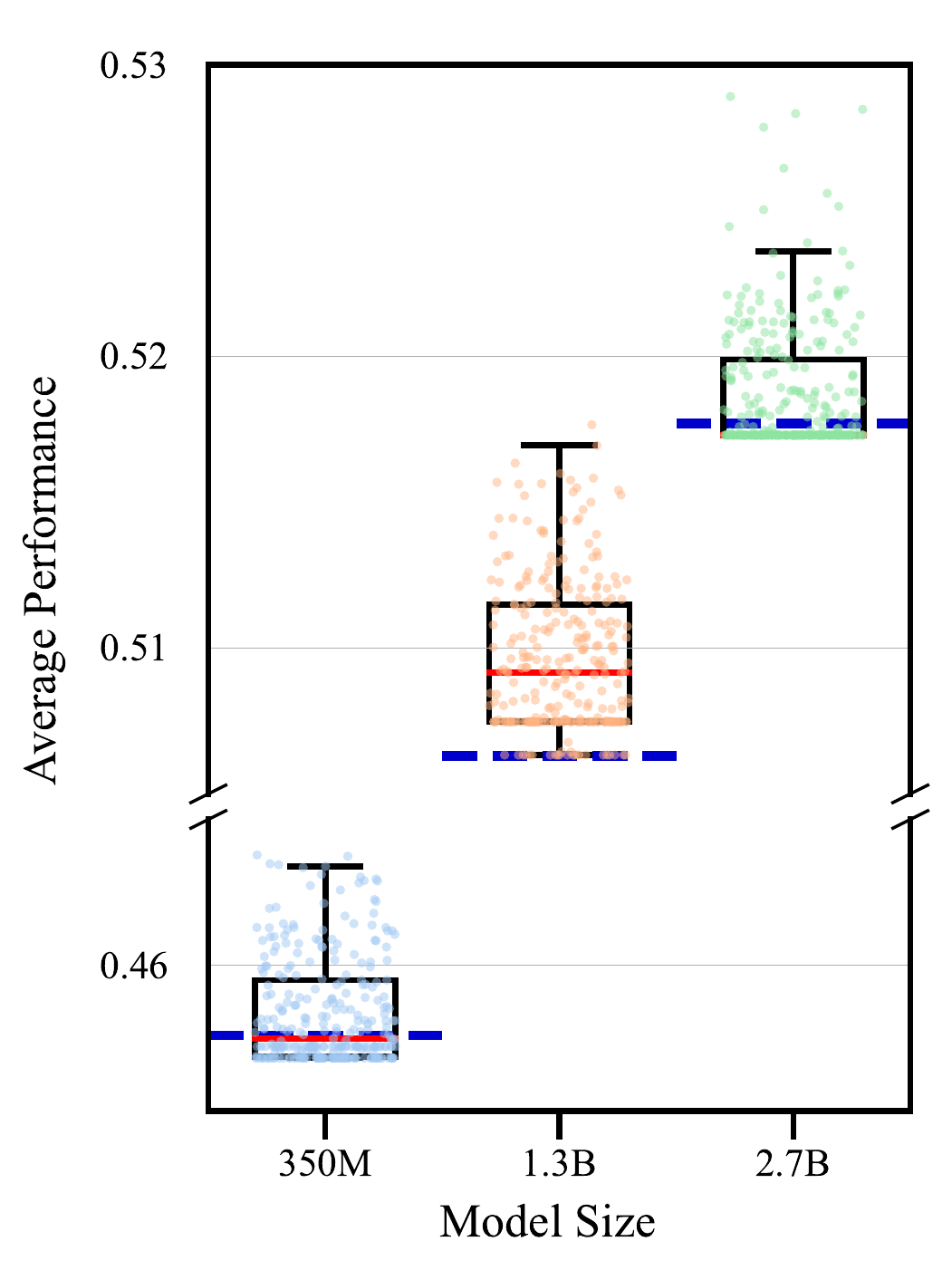}
\centering
\caption{Average validation accuracy measured on 8 classification tasks. A single dot represents a single GAP run, total of 300 runs per LM size. The dashed horizontal lines indicate performance of baseline LMs with same size.}
\label{fig:val_all}
\end{figure}

\subsection{Classification Tasks}
The average validation performances of the 8 classification tasks when performing \method on the OPT LMs are shown in Figure \ref{fig:val_all}.
While \method fails to provide consistent improvements for 350M LMs and 2.7B LMs, mostly resulting in a degradation of performance as shown by the median performance underperforming the baselines, the LMs show considerable performance gains in some cases for the larger LMs. This result suggests that although GAP does not show steady improvement of generalization for the classification tasks unlike the dialogue tasks, it does show some potential for improvement considering that some runs did result in substantial improvements. We leave choosing the right text samples to perform \method on for a consistent performance enhancement on classification tasks for future work.

\begin{figure}[t!]
\includegraphics[width=1\linewidth]{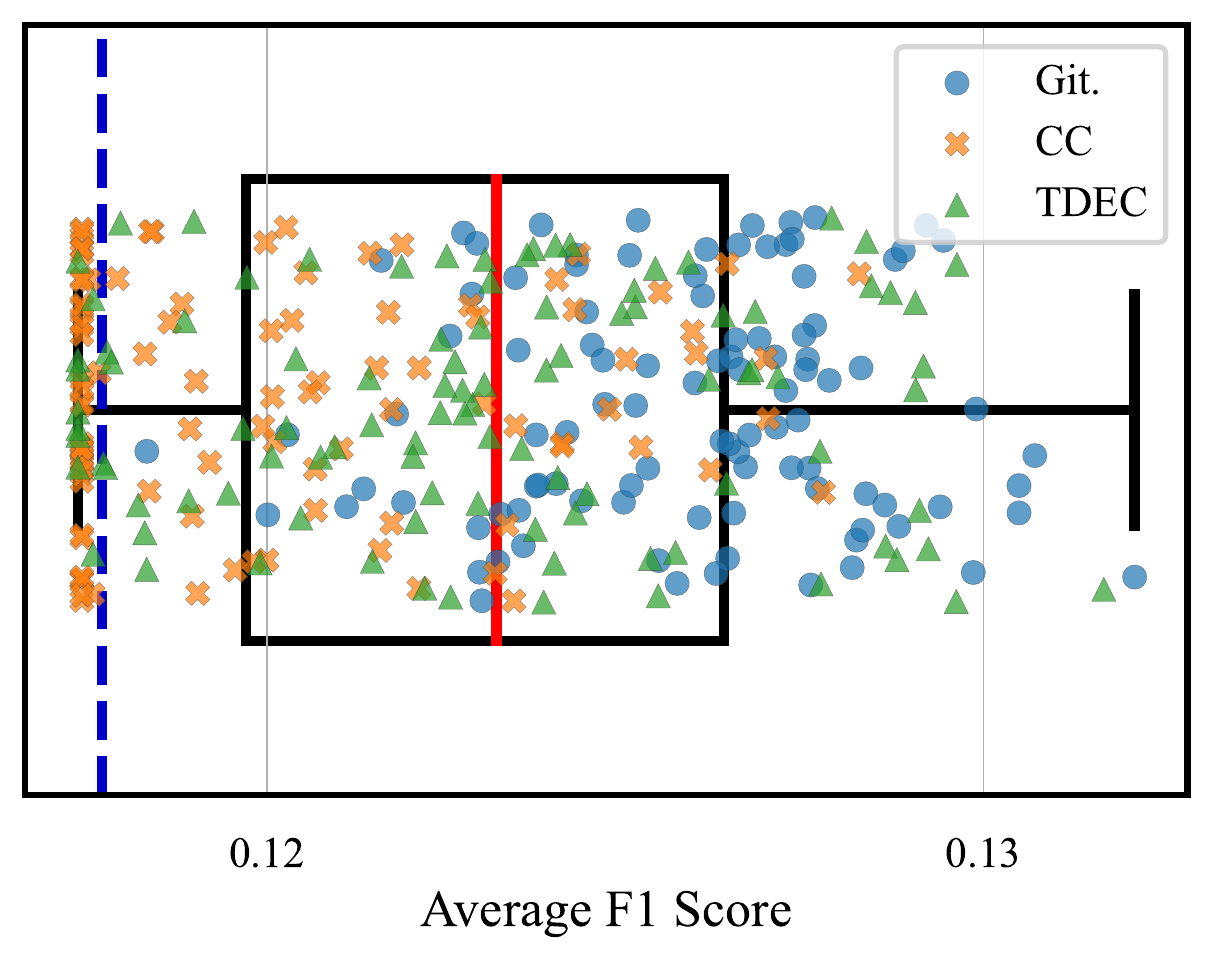}
\centering
\caption{Average validation F1 score improvements for 350M-LMs measured on four dialogue datasets. Each symbol represents a single GAP run, with 100 runs per corpus and a total of 300 runs. The dashed vertical line indicates the performance of 350M-OPT LM.}
\label{fig:analysis_2.7B}
\end{figure}

\begin{table}[t]
\centering
\begin{tabularx}{0.4\textwidth}{l|c|ccc}
\toprule
\textbf{Model} & \textbf{All} & \textbf{Git.} & \textbf{CC} & \textbf{TDEC}\\
\midrule
350M & 12.3 & \textbf{12.6} & 11.9 & 12.3 \\
\midrule
1.3B & 13.7 & \textbf{13.8} & 13.6 & 13.5 \\
\midrule
2.7B & 14.1 & \textbf{14.3} & 14.2 & 13.9 \\
\bottomrule
\end{tabularx}
\caption{
Average validation F1 score measured on four dialogue datasets, split into the origin of the unlabeled data. The values for \textbf{Git.}, \textbf{CC}, \textbf{TDEC} are the median value of the 100 runs for each corpus. The value for \textbf{All} is the median value of the 300 total GAP runs.
}
\label{table:analysis_quant}
\end{table}

\subsection{Analysis of \method}
Figure \ref{fig:analysis_2.7B} shows the average performance of the 300 \method runs for the 350M LMs (zoomed-in version of Figure \ref{fig:dialog_val_f1}). To observe the effect of LMs' familiarity to the unlabeled data, we plot the dots with different symbols with respect to the corpus. Interestingly,  samples from the unfamiliar corpus (Github) results in significant improvements, mostly achieving higher scores than the median score. Consistent findings are also evident in Table \ref{table:analysis_quant}, with Github achieving the highest median F1 scores across all model sizes. This suggests that future applications of \method can be applied more efficiently by mostly using unfamiliar (out-of-distribution) text. Additional figures for other LM sizes are available in Appendix C.

\section{Conclusion}
In this work, we introduce \method, a novel method of improving the generalization capability of LMs without any task-specifc data by sampling random text and performing gradient ascent for a few steps. We show that our approach is (1) simple to use, (2) effective in making more robust LMs, and (3) has much room for improvements for future work when scaling the number of \method runs (e.g. >300) and choosing specific text samples (e.g. out-of-distribution text) to perform \method on. Thus, we urge the community to consider \method when prompting off-the-shelf pretrained LMs for performing diverse downstream NLP tasks. 

\section*{Limitations}
While we show that applying \method can result in a significant improvement in the generalization capability of LMs, especially for dialogue tasks, we are only able to show 300 \method runs for each LM size in  this work. We leave scaling the number of \method runs, as well as selecting \textit{specific} text samples to perform \method on for future work. Furthermore, a separate validation set of the tasks at interest are needed in order to choose the best checkpoint when performing \method. Future work may look for other task-agonostic cues such as language modeling loss to determine the best checkpoint to use for inference. 

\section*{Acknowledgements}
This work was partly supported by KAIST-NAVER Hypercreative AI Center (80\%) and Institute of Information \& communications Technology Planning \& Evaluation (IITP) grant funded by the Korea government (MSIT) (No.2022-0-00113, Developing a Sustainable Collaborative Multi-modal Lifelong Learning Framework, 20\%).
\bibliography{anthology,custom}
\bibliographystyle{acl_natbib}

\appendix

\section{Task Prompts}
\label{appen:task_prompt}
Table \ref{table:prompts} shows the prompts we use for each of the 12 benchmark dataset to enable zero-shot/few-shot learning. For dialogue tasks (Wizard of Wikipedia, Blended Skill Talks, Empathetic Dialogues, WizInt), we use the prompts used by \citet{zhang2022opt}.

\section{Details of Experimental Configurations}
\label{appen:configs}
In this section, we give further details of our main experimental setting of performing \method. We use Adam optimizer~\cite{Kingma2014adam} with a constant learning rate of 5e-5 with no weight decay and no dropout. 

For the dialogue tasks, we adopt the settings of ~\citet{zhang2022opt} and prompt the LM with alternating "User 1:" and "User 2:" lines of dialogue (examples shown in Appendix \ref{appen:task_prompt}). To generate tokens, we employ greedy decoding method and set a maximum generation length of 32 tokens. For the classification tasks, we use a \textit{verbalizer} method by selecting the output option with higher log-likelihood following \citet{brown2020language, sanh2021multitask}. We use unigram F1 score as our main metric for the dialogue generation tasks and accuracy for the classification tasks.

For the diverse metrics used for evaluation on the test sets of the 4 dialogue tasks, MAUVE~\cite{pillutla-etal:mauve:neurips2021} compares the text representation of the LM generated-response to human-written text, higher values indicate greater similarity to human-written text. Diversity metric~\cite{su2022a} measures token-level repetition, with higher values indicating greater diversity and less repetition in the generated text. 

\section{Full Results}
\label{appen:full_results}
Tables \ref{table:val_dialog_f1_median} and \ref{table:val_class_acc_median} show the median validation score of all 300 \method runs. For classification tasks, the median values do not show significant improvements. However for dialogue tasks, \method shows considerable improvements across all tasks.

Tables \ref{table:test_dialog_f1}, \ref{table:test_dialog_mauve}, \ref{table:test_dialog_diversity} and \ref{table:test_dialog_length} show the individual test performance for each dialogue dataset. The four dialogue datasets are: Blended Skill Talks (\textbf{BST}), Empathetic Dialogues (\textbf{ED}), Wizard of Wikipedia (\textbf{WoW}) and \textbf{WizInt}. Our models demonstrate superior performance compared to their same sized baselines on every metrics in all four task.

\begin{figure}[ht!]
\includegraphics[width=1\linewidth]{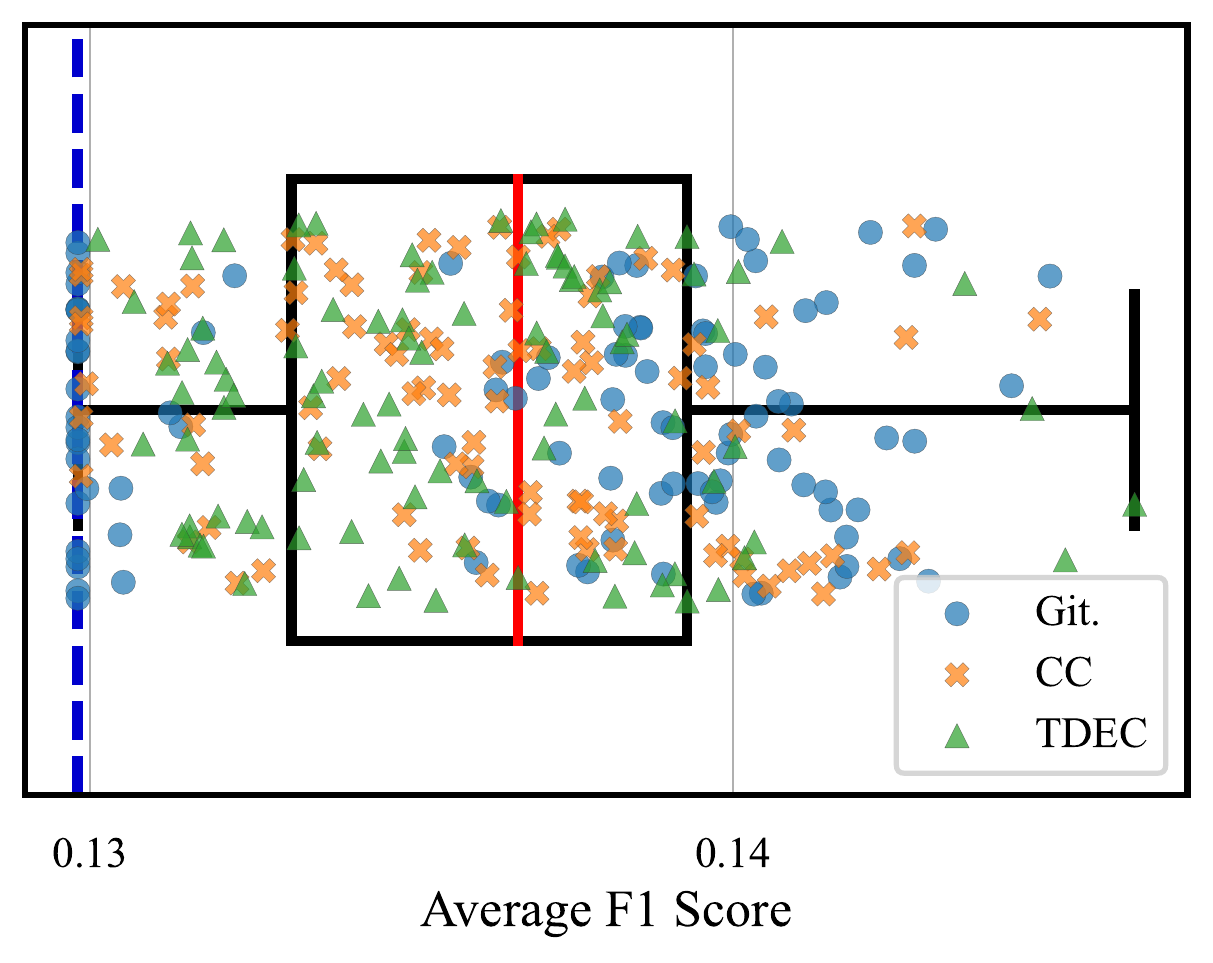}
\centering
\caption{Analysis of average validation F1 score improvements for 1.3B-LMs measured on four dialogue datasets.}
\label{fig:analysis_350M}
\end{figure}

\begin{figure}[ht!]
\includegraphics[width=1\linewidth]{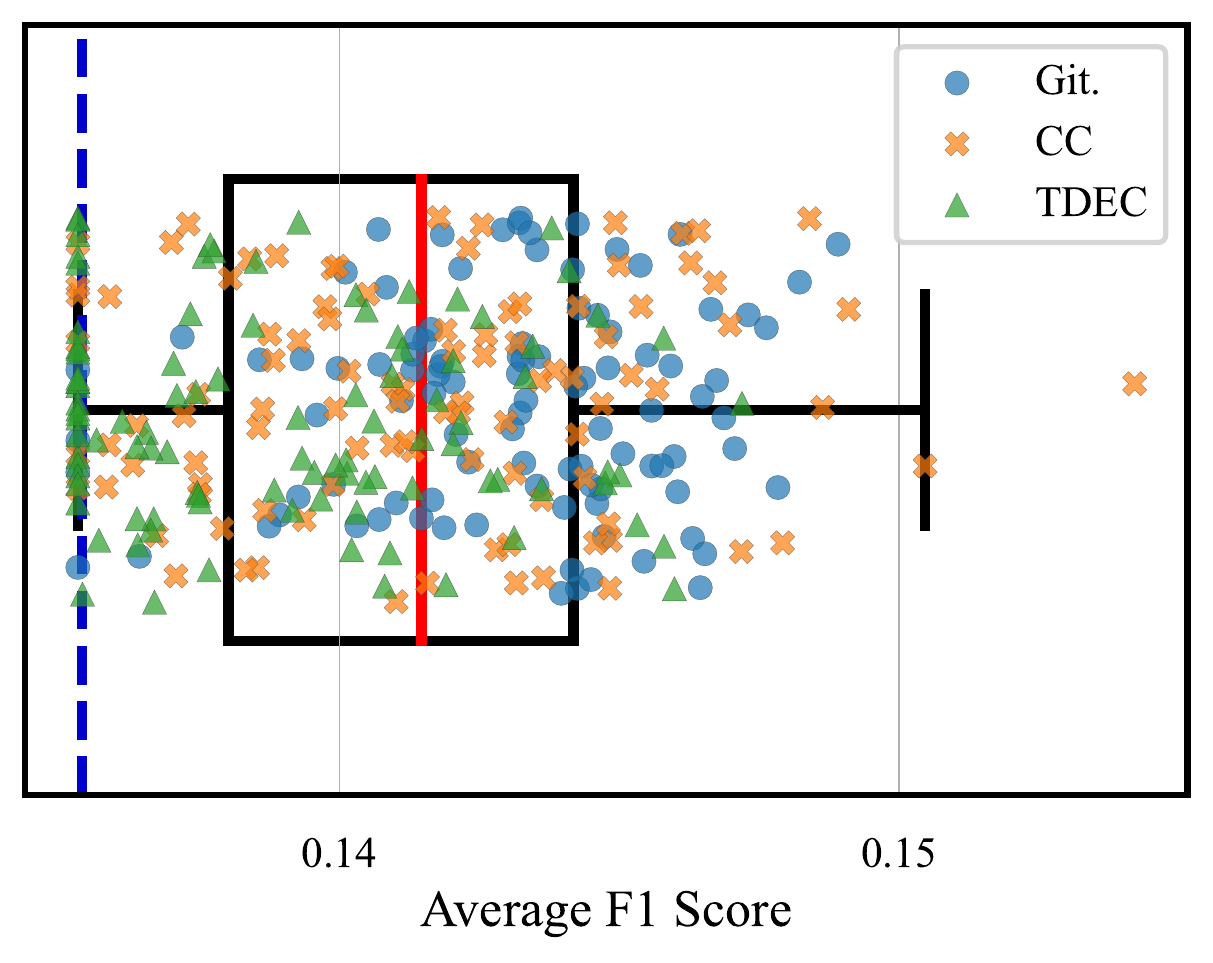}
\centering
\caption{Analysis of average validation F1 score improvements for 2.7B-LMs measured on four dialogue datasets.}
\label{fig:analysis_1.3B}
\end{figure}

Figures \ref{fig:analysis_350M} and \ref{fig:analysis_1.3B} represent the familiarity analysis results for 1.3B and 2.7B sized models, respectively. For both 1.3B and 2.7B models, data sampled from the out-of-domain corpora (Github) results in reliable performance gains. For the bigger sized models, in-domain corpora (CC) also results in competitive performance gains, suggesting larger sized morels are more robust to \method data selection.

\section{Examples of Random Data}
\label{appen:random_data}
Table \ref{table:data_example} shows examples of the random data we use to apply GAP to OPT LMs. Specifically, they are the best performing data for each model size.

\section{Examples of Dialogue Generation Outputs}
\label{appen:examples}

Table \ref{table:generation_example} shows some examples of text generated by baseline models and our models trained with \method. Notice that our models generate diverse and interesting text while also maintaining coherence to the given dialogue history.

\section{Details of Human Evaluation}
\label{sec:human_eval_detail}

\begin{figure*}[hbtp]
\includegraphics[width=1\linewidth]{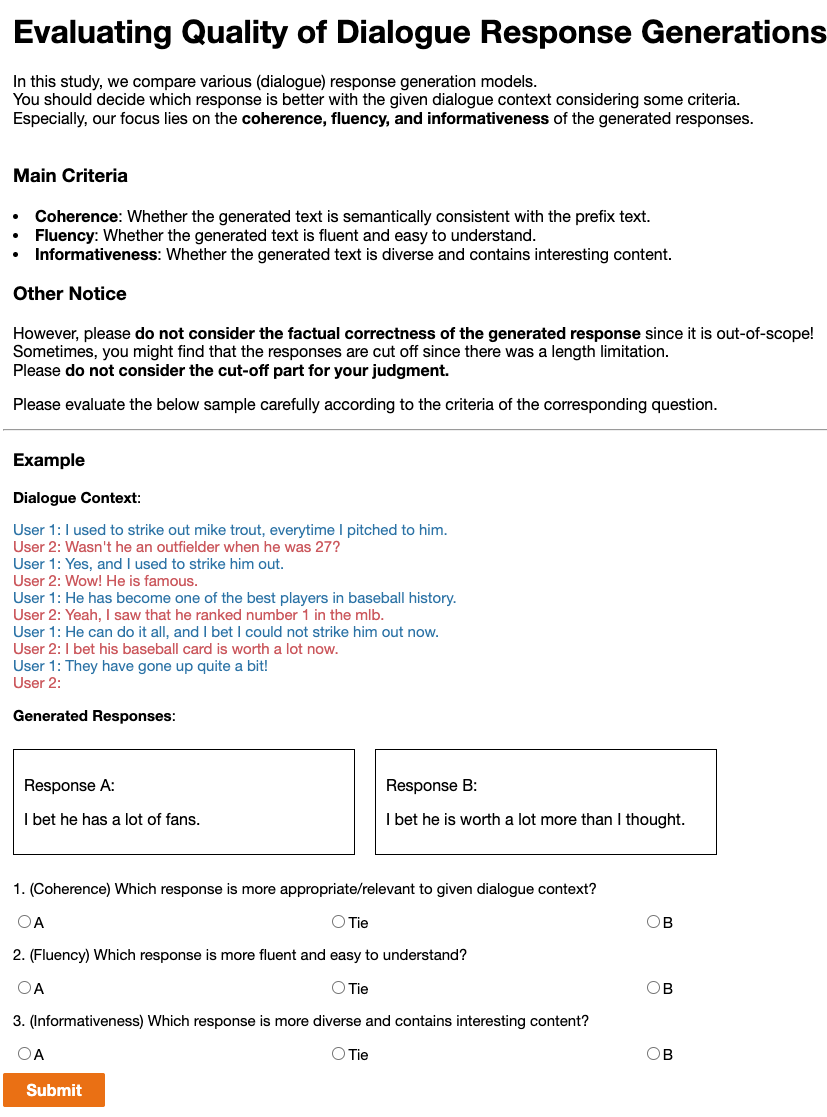}
\centering
\caption{An example of the Mturk interface used for the human evaluation of the dialogue response generation quality.}
\label{fig:human_eval}
\end{figure*}

We conduct the human evaluation on Amazon Mechanical Turk (AMT). An example of the interface shown to the workers is shown in Figure \ref{fig:human_eval}. Specifically, we recruit three different annotators for each comparison pair with a compensation of 1\$ per instance. We include brief instructions on the evaluation including descriptions of three metrics. Then, we ask the workers to compare each generated (or ground-truth for human baseline) response pair with the given dialogue context. We evaluate 200 samples in total, including 100 for the OPT baseline and 100 for the human upper bounds. The Fleiss kappa among the workers is calculated as 0.36, which indicates moderate-level agreements. We also test the significance between the comparing systems via a bootstrap test with 100,000 samplings.

\begin{table*}[t]
\centering
\fontsize{8}{10}\selectfont
\caption{\small Full list of the prompts used for the 12 evaluation datasets.}
\begin{tabularx}{1\textwidth}{>{\hsize=.6\hsize\centering\arraybackslash}X|>{\raggedright\arraybackslash}X}
    \toprule
    \textbf{Dataset} & \textbf{Prompt} \\
    \midrule
    \textbf{PIQA} &  \{goal\} [option] \\
    \midrule
    \textbf{ARC-Easy/Challenge} & \{question\} [option] \\
    \midrule
    \textbf{COPA} & \{premise\} [option] \\
    \midrule
    \textbf{HellaSwag} & \{input\} [option] \\
    \midrule 
    \textbf{Winogrande} & \{sentence\} [option] \\
    \midrule 
    \textbf{MathQA} & \{problem\} [option] \\
    \midrule 
    \textbf{PubmedQA} & Question: \{problem\} \textbackslash nAnswer: [option] \\
    \midrule 
    \textbf{Wizard of Wikipedia}, \textbf{Blended Skill Talks}, \textbf{Empathetic Dialogues}, \textbf{WizInt} & User 1: \{turn\}\textbackslash nUser 2: \{turn\}\textbackslash nUser 1: \{turn\}\textbackslash n ... User 2: \\ 
    \bottomrule
\end{tabularx}
\label{table:prompts}
\end{table*}

\begin{table*}[h!]
\centering
\begin{tabularx}{0.60\textwidth}{l|>{\centering\arraybackslash}X|>{\centering\arraybackslash}X>{\centering\arraybackslash}X>{\centering\arraybackslash}X>{\centering\arraybackslash}X}
\toprule
\textbf{Model} & \textbf{Avg.} & \textbf{BST} & \textbf{ED} & \textbf{WoW} & \textbf{WizInt}\\
\midrule
350M & 11.77 & 11.88 & 10.17 & 12.05 & 13.00\\ 
\textit{+ GAP} & \textbf{12.31} & \textbf{12.45} & \textbf{10.64} & \textbf{12.37} & \textbf{13.78}  \\
\midrule
1.3B & 12.98 & 14.04 & 12.35 & 11.68 & 13.85 \\
\textit{+ GAP} & \textbf{13.60} & \textbf{14.45} & \textbf{12.58} & \textbf{12.37} &\textbf{15.02} \\
\midrule
2.7B & 13.54 & 13.18 & 12.42 & 12.86 & \textbf{15.69} \\
\textit{+ GAP} & \textbf{14.09} & \textbf{13.90} & \textbf{13.03} & \textbf{13.76} & 15.65 \\
\midrule
6.7B & 14.51 & 14.93 & 13.71 & 14.24 & 15.18 \\
\bottomrule
\end{tabularx}
\caption{
\textbf{Validation F1-score} of OPT baselines and median \textbf{validation F1-score} of all GAP runs, measured on four dialogue datasets: Blended Skill Talks (\textbf{BST}), Empathetic Dialogues (\textbf{ED}), Wizard of Wikipedia (\textbf{WoW}) and \textbf{WizInt}.}
\label{table:val_dialog_f1_median}
\end{table*}

\begin{table*}[h!]
\centering
\begin{tabularx}{1\textwidth}{l|>{\centering\arraybackslash}X|>{\centering\arraybackslash}X>{\centering\arraybackslash}X>{\centering\arraybackslash}X>{\centering\arraybackslash}X>{\centering\arraybackslash}X>{\centering\arraybackslash}X>{\centering\arraybackslash}X>{\centering\arraybackslash}X}
\toprule
\textbf{Model} & \textbf{Avg.} & \textbf{ARC-Chall.} & \textbf{ARC-Easy} & \textbf{Hella-swag} & \textbf{MathQA} & \textbf{PIQA} & \textbf{Pubmed-QA} & \textbf{COPA} & \textbf{Wino-grande}\\
\midrule
350M & 45.76 & 11.64 & 45.63 & 35.94 & \textbf{21.88} & 67.50 & \textbf{54.37} & 69.00 & 53.13\\ 
\textit{+ GAP} & \textbf{45.84} & \textbf{19.32} & 45.63 & \textbf{36.88} & 21.25 & 67.50 & 53.75 & 69.00 & \textbf{53.44}  \\
\midrule
1.3B & 50.63 & 24.07 & 56.25 & 39.38 & 22.81 & 69.38 & 58.44 & 76.00 & 58.75 \\
\textit{+ GAP} & \textbf{50.91} & \textbf{24.75} & 56.25 & \textbf{40.00} &\textbf{23.13} & \textbf{70.00} & 58.44 & 76.00 & 58.75 \\
\midrule
2.7B & \textbf{51.77} & 26.78 & 57.50 & 41.87 & 21.25 & \textbf{72.50} & 58.44 & 78.00 & 57.81 \\
\textit{+ GAP} & 51.73 & 26.78 & 57.50 & 41.87 & 21.25 & 72.19 & 58.44 & 78.00 & 57.81 \\
\midrule
6.7B & 54.39 & 32.20 & 61.87 & 45.63 & 21.25 & 75.94 & 58.44 & 77.00 & 62.81 \\
\bottomrule
\end{tabularx}
\caption{
\textbf{Validation accuracy} of OPT baselines and median \textbf{validation accuracy} of all GAP runs, measured on classification datasets.}
\label{table:val_class_acc_median}
\end{table*}

\begin{table}[ht!]
\centering
\begin{tabularx}{0.44\textwidth}{l|ccccc}
\toprule
\textbf{Model} & \textbf{BST} & \textbf{ED} & \textbf{WoW} & \textbf{WizInt}\\
\midrule
350M & 11.18 & 10.43 & 13.24 & 10.92 \\
\textit{+ GAP} & \textbf{12.68} & \textbf{11.38} & \textbf{13.89} & \textbf{12.13} \\
\midrule
1.3B & 14.26 & 12.51 & 14.38 & 13.01 \\
\textit{+ GAP} & \textbf{14.83} & \textbf{12.74} & \textbf{15.18} & \textbf{13.37} \\
\midrule
2.7B & 14.00 & 13.09 & 14.40 & 13.58 \\
\textit{+ GAP} & \textbf{15.12} & \textbf{13.71} & \textbf{15.40} & \textbf{14.45} \\
\midrule
6.7B & 15.04 & 13.79 & 15.19 & 13.92 \\
\bottomrule
\end{tabularx}
\caption{
Test \textbf{F1-score} of our best performing GAP models and OPT baselines on each dialogue datasets.}
\label{table:test_dialog_f1}
\end{table}

\begin{table}[ht!]
\centering
\begin{tabularx}{0.44\textwidth}{l|ccccc}
\toprule
\textbf{Model} & \textbf{BST} & \textbf{ED} & \textbf{WoW} & \textbf{WizInt}\\
\midrule
350M & 48.73 & 31.01 & 53.58 & 43.91 \\
\textit{+ GAP} & \textbf{74.87} & \textbf{62.29} & \textbf{82.37} & \textbf{82.55} \\
\midrule
1.3B & 52.6 & 53.0 & 40.8 & 46.2 \\
\textit{+ GAP} & \textbf{74.7} & \textbf{54.5} & \textbf{76.4} & \textbf{72.44} \\
\midrule
2.7B & 59.8 & 49.4 & 55.4 & 40.6 \\
\textit{+ GAP} & \textbf{82.2} & \textbf{51.3} & \textbf{86.7} & \textbf{71.5} \\
\midrule
6.7B & 55.7 & 43.4 & 56.3 & 48.8 \\
\bottomrule
\end{tabularx}
\caption{
Test \textbf{MAUVE} of our best performing GAP models and OPT baselines on each dialogue datasets.}
\label{table:test_dialog_mauve}
\end{table}

\begin{table}[ht!]
\centering
\begin{tabularx}{0.44\textwidth}{l|ccccc}
\toprule
\textbf{Model} & \textbf{BST} & \textbf{ED} & \textbf{WoW} & \textbf{WizInt}\\
\midrule
350M & 69.29 & 85.01 & 62.64 & 79.34 \\
\textit{+ GAP} & \textbf{83.22} & \textbf{91.79} & \textbf{82.96} & \textbf{91.09} \\
\midrule
1.3B & 82.62 & 84.43 & 81.07 & 83.23 \\
\textit{+ GAP} & \textbf{86.78} & \textbf{88.99} & \textbf{84.33} & \textbf{86.64} \\
\midrule
2.7B & 85.36 & 91.09 & 82.04 & 89.26 \\
\textit{+ GAP} & \textbf{93.99} & \textbf{96.22} & \textbf{89.73} & \textbf{92.38} \\
\midrule
6.7B & 86.95 & 92.29 & 81.28 & 92.67 \\
\bottomrule
\end{tabularx}
\caption{
Test \textbf{diversity} of our best performing GAP models and OPT baselines on each dialogue datasets.}
\label{table:test_dialog_diversity}
\end{table}

\begin{table}[ht!]
\centering
\begin{tabularx}{0.44\textwidth}{l|ccccc}
\toprule
\textbf{Model} & \textbf{BST} & \textbf{ED} & \textbf{WoW} & \textbf{WizInt}\\
\midrule
350M & 10.91 & 10.65 & 13.4 & 12.23 \\
\textit{+ GAP} & \textbf{13.23} & \textbf{13.26} & \textbf{15.86} & \textbf{15.35} \\
\midrule
1.3B & 10.69 & 11.18 & 11.95 & 11.72 \\
\textit{+ GAP} & \textbf{12.89} & \textbf{12.49} & \textbf{15.05} & \textbf{14.8} \\
\midrule
2.7B & 10.4 & 10.72 & 12.39 & 11.58 \\
\textit{+ GAP} & \textbf{13.09} & \textbf{13.98} & \textbf{15.83} & \textbf{15.21} \\
\midrule
6.7B & 11.25 & 10.89 & 13.36 & 12.22 \\
\bottomrule
\end{tabularx}
\caption{
Test \textbf{generation length} of our best performing GAP models and OPT baselines on each dialogue datasets.}
\label{table:test_dialog_length}
\end{table}

\begin{table*}[hbtp]
\centering
\begin{tabular}{c | p{0.8\linewidth}}
\toprule
\textbf{Model} & \textbf{Text}\\
\toprule
350M + \textit{GAP} &    "metadata": {},\textbackslash n   "source": [\textbackslash n    "Canary rollouts are used to release new models safely to only a small subset of users such as 5\%. They are useful if you want to test in live production without affecting the entire user base. Since the majority of traffic goes to the existing model, the cluster size of the canary model can be relatively small since it’s only receiving 5\% traffic."\textbackslash n   ]\textbackslash n  \},\textbackslash n  \{\textbackslash n   "cell\_type": "markdown",\textbackslash n   "metadata": \{\},\textbackslash n   "source": [\textbackslash n    "Instead of `deploy()`, we can create an `Endpoint Configuration` with multiple variants for canary rollouts and A/B testing."\textbackslash n   ]\textbackslash n  \},\textbackslash n  \{\textbackslash n   "cell\_type": "code",\textbackslash n   "execution\_count": null,\textbackslash n   \\
\midrule
1.3B + \textit{GAP} & \textbackslash n\textbackslash tld d, 08\textbackslash n\textbackslash tld a, 10\textbackslash n\textbackslash tld b, 11\textbackslash nlfill\_vram:\textbackslash n\textbackslash tld(hl++), a\textbackslash n\textbackslash tadd a, b\textbackslash n\textbackslash tdec d\textbackslash n\textbackslash tjrnz lfill\_vram\textbackslash n\textbackslash tld a, 90\textbackslash n\textbackslash tldff(45), a\textbackslash n\textbackslash tld a, 40\textbackslash n\textbackslash tldff(41), a\textbackslash n\textbackslash txor a, a\textbackslash n\textbackslash tldff(0f), a\textbackslash n\textbackslash tld a, 02\textbackslash n\textbackslash tldff(ff), a\textbackslash n\textbackslash tei\textbackslash n\textbackslash thalt\textbackslash n\textbackslash n.text@7000\textbackslash nlprint4:\textbackslash n\textbackslash tld b, 90\textbackslash n\textbackslash tcall lwaitly\_b\textbackslash n\textbackslash txor a, a\textbackslash n\textbackslash tldff(40), a\textbackslash n\textbackslash tld bc, 7a00\textbackslash n\textbackslash tld hl, 8000\textbackslash n\textbackslash tld d, 00\textbackslash nlprint\_copytiles:\textbackslash n\textbackslash tld a, (bc)\textbackslash n\textbackslash tinc bc \\
\midrule
2.7B + \textit{GAP} &  crafts of Jharkhand. The people of the state who belong to the different ethnic groups in the state are mainly engaged in this form of craft.\textbackslash n\textbackslash nThe Jharkhand bamboo crafts that the tribal people of Jharkhand are engaged in show a great deal of intricate and fascinating hand work, which is quite unique to the state of India. The major articles that are made out of bamboo in Jharkhand include baskets and accessories used for fishing and hunting.\textbackslash n\textbackslash nThe bamboo crafts in Jharkhand that the ethnic people of the state of Jharkhand make are mostly carved out of the bamboos available locally. The variety of bamboo produced by the bamboo grooves of the state is not very thick. However, these bamboos are suitable for the different kinds of bamboo crafts at Jharkhand, since they are not weak and yet can be twisted and turned to a great extent.\textbackslash n\textbackslash nMetal Works of Jharkhand\textbackslash n\textbackslash nMetal \\
\bottomrule
\end{tabular}
\caption{
Example of the best performing random data for each model size.
}
\label{table:data_example}
\end{table*}

\begin{table*}[hbtp]
\centering
\begin{tabular}{c | p{0.8\linewidth}}
\toprule
\textbf{Model} & \textbf{Generation}\\
\toprule
\multicolumn{2}{p{0.2\linewidth}}{
\makecell[l]{
User 1: I know absolutely nothing about archery, but would be interested in your views on it.\\User 2: It is really very fun. It can be considered a sport or a skill where a bow is used to\\propel arrows.\\User 1: Hmm. Do you go to classes? It sound like such fun - I've seen it in movies but not\\really thought much about it.\\User 2: I do not. It used to be mainly for hunting and fighting, but not anymore.\\User 1: Perhaps I should give it a go, not for hunting but for sport.My husband does bow hunt,\\if that's a similar sort of thing. He seems to be quite good at it.\\User 2: It is an increasingly popular competitive sport and recreational activity.\\User 1: Interesting. Do many countries participate? It would be fun to see the ways countries\\differentiate.\\User 2:}} \\
\midrule
1.3B & There are many countries that participate. \\
\midrule
\textit{+ GAP} & There are many countries that participate in the sport. Some countries have national competitions, while others have regional competitions. \\
\midrule
\midrule
\multicolumn{2}{p{0.2\linewidth}}{
\makecell[l]{
User 1: My friend bought a big house and I'm happy for her but at the same time I'm a little\\bit jealous.\\User 2: Big houses are a pain in the neck to keep clean.\\User 1: True, but there is so much space for her kid to run and play at and I live in a\\tiny apartment with my son.\\User 2:}} \\
\midrule
2.7B & I'm sure she'll be happy with her new house. \\
\midrule
\textit{+ GAP} & I know what you mean. My house is so small that I can't even fit my son's toys in the living room. \\
\midrule
\midrule
\multicolumn{2}{p{0.2\linewidth}}{
\makecell[l]{
 User 1: I am an accountant. What is your profession.\\
 User 2: pacioli established accounting in 1494 if I remember correctly ha. I work in healthcare.\\
User 1: What is your role in healthcare. I have been an accountant for 5 years. \\
User 2: I have an administrative role at a non-profit hospital. \\
User 1: That is interesting. What other things will you like to tell me about your profession. \\
User 2: I work in obtaining funding for the hospital. What type of accounting do you do \\
User 1: I do general accounting. \\
User 2: Lee had major impacts in the field of cost accounting. \\
User 1: That is interesting to know. Who is lee. \\
User 2:}} \\
\midrule
2.7B & Lee was a pioneer in cost accounting. \\
\midrule
\textit{+ GAP} & Lee was a famous American accountant. He was the founder of the American Institute of Certified Public Accountants (AICPA). \\
\bottomrule
\end{tabular}
\caption{
Examples of texts generated by baseline OPT models and our GAP applied models, given dialogue histories as prompts.
}
\label{table:generation_example}
\end{table*}

\end{document}